\documentclass[twocolumn, switch]{article} 
\usepackage{silence}
\usepackage{graphicx}
\usepackage[numbers, sort&compress]{natbib}
\usepackage{doi}
\usepackage[section]{placeins}
\usepackage{tabularx}
\usepackage{multirow}
\usepackage{makecell}
\usepackage{lineno}
\usepackage{xcolor}
\newcommand{\best}[1]{\textcolor{red}{\textbf{#1}}}
\newcommand{\secondbest}[1]{\textcolor{blue}{#1}}
\usepackage{hyperref}
\hypersetup{
    colorlinks=true,
    linkcolor=cyan,
    urlcolor=cyan,
    citecolor=cyan,
}
\usepackage[font=small, labelfont=bf, labelsep=period]{caption}

\usepackage{preprint}
\usepackage{algorithm}
\usepackage{algorithmic}
\usepackage{amsmath, amsthm, amssymb, amsfonts}

\usepackage{booktabs} 		
\usepackage{nicefrac}		
\usepackage{microtype}		
\usepackage{lineno}		
\usepackage{float}			

\usepackage{lipsum}		

\usepackage{titlesec}
\titleformat{\section}
  {\large\bfseries\scshape}
  {\thesection}
  {1em}
  {}

\titleformat{\subsection}
  {\normalsize\bfseries}
  {\thesubsection}
  {1em}
  {}

\titlespacing*{\section}{0pt}{1.2em}{0.5em}
\titlespacing*{\subsection}{0pt}{0.8em}{0.3em}
\titlespacing*{\subsubsection}{0pt}{0.5em}{0.2em}

\title{DB-KAUNet: An Adaptive Dual Branch Kolmogorov-Arnold UNet for Retinal Vessel Segmentation}
\usepackage{titling}
\usepackage{orcidlink}
\usepackage{footmisc}
\newcommand{\Author}[2]{%
  \textbf{#1}\textsuperscript{#2} 
}

\author{
  \Author{Hongyu Xu}{1} \and
  \Author{Panpan Meng}{2} \and
  \Author{Meng Wang}{3,4} \and
  \Author{Dayu Hu}{5} \and
  \Author{Liming Liang}{6} \and
  \Author{Xiaoqi Sheng}{7,*}
}

\date{%
  \footnotesize
  \textsuperscript{1}School of Computer Science and Software Engineering, Southwest University, Chongqing 400715, China\\
  \textsuperscript{2}Innovation Centre of Ministry of Education for Development and Diseases, the Sixth Affiliated Hospital, School of Medicine, South China University of Technology, Guangzhou 511442, China\\
  \textsuperscript{3}Centre for Innovation and Precision Eye Health, Yong Loo Lin School of Medicine, National University of Singapore, Singapore 119228, Singapore\\
  \textsuperscript{4}Department of Ophthalmology, Yong Loo Lin School of Medicine, National University of Singapore, Singapore 119228, Singapore\\
  \textsuperscript{5}College of Medicine and Biological Information Engineering, Northeastern University, Shenyang 110169, China\\
  \textsuperscript{6}School of Electrical Engineering Automation, Jiangxi University of Science and Technology, Ganzhou 341000, China\\
  \textsuperscript{7}School of Future Technology, South China University of Technology, Guangzhou 511442, China\\[1em]
}

\usepackage{fancyhdr}
\newcommand\blfootnote[1]{%
  \begingroup
  \renewcommand\thefootnote{}\footnote{#1}%
  \addtocounter{footnote}{-1}%
  \endgroup
}
\begin{document}

\twocolumn[ 
  \begin{@twocolumnfalse} 

\maketitle
\thispagestyle{fancy}

\begin{abstract}
Accurate segmentation of retinal vessels is crucial for the clinical diagnosis of numerous ophthalmic and systemic diseases. However, traditional Convolutional Neural Network (CNN) methods exhibit inherent limitations, struggling to capture long-range dependencies and complex nonlinear relationships. To address the above limitations, an Adaptive Dual Branch Kolmogorov-Arnold UNet (DB-KAUNet) is proposed for retinal vessel segmentation. In DB-KAUNet, we design a Heterogeneous Dual-Branch Encoder (HDBE) that features parallel CNN and Transformer pathways. The HDBE strategically interleaves standard CNN and Transformer blocks with novel KANConv and KAT blocks, enabling the model to form a comprehensive feature representation. To optimize feature processing, we integrate several critical components into the HDBE. First, a Cross-Branch Channel Interaction (CCI) module is embedded to facilitate efficient interaction of channel features between the parallel pathways. Second, an attention-based Spatial Feature Enhancement (SFE) module is employed to enhance spatial features and fuse the outputs from both branches. Building upon the SFE module, an advanced Spatial Feature Enhancement with Geometrically Adaptive Fusion (SFE-GAF) module is subsequently developed. In the SFE-GAF module, adaptive sampling is utilized to focus on true vessel morphology precisely. The adaptive process strengthens salient vascular features while significantly reducing background noise and computational overhead. Extensive experiments on the DRIVE, STARE, and CHASE\_DB1 datasets validate that DB-KAUNet achieves leading segmentation performance and demonstrates exceptional robustness.
\end{abstract}
\keywords{Retinal Vessel Segmentation \and Kolmogorov-Arnold Networks \and Hybrid CNN-Transformer \and Dual Branch Architecture} 
\vspace{0.7cm}

  \end{@twocolumnfalse} 
] 

\blfootnote{\textbf{*}Corresponding author\\ \noindent 
\textit{Email address:} \texttt{xqsheng@scut.edu.cn} (Xiaoqi Sheng)
}

\section{Introduction}
\label{sec1}
Analyzing retinal vessel morphology and topology is clinically vital for diagnosing and managing ocular diseases. For instance, Diabetic Retinopathy (DR) is characterized by vessel swelling due to leakage and obstruction \cite{Carol2011Retinal}. In contrast, Hypertensive Retinopathy (HR) is indicated by increased vessel tortuosity or narrowing \cite{2014Classification}. Glaucoma presents a different pattern, often involving arterial narrowing, reduced tortuosity, and microvascular dropout \cite{2017Retinal}. Therefore, accurately segmenting vessels from fundus images has become a clinically urgent task. Vessel segmentation methods are categorized as manual or automatic. Although manual segmentation is performed by experts, the procedure is laborious, time-consuming, and tedious. Furthermore, the accuracy of the segmentation results depends heavily on the expertise of the individual operator. Consequently, automatic methods have been developed to overcome the limitations inherent in the manual approach.

\begin{figure}[!htb]
\centering
\includegraphics[width=1.0\columnwidth]{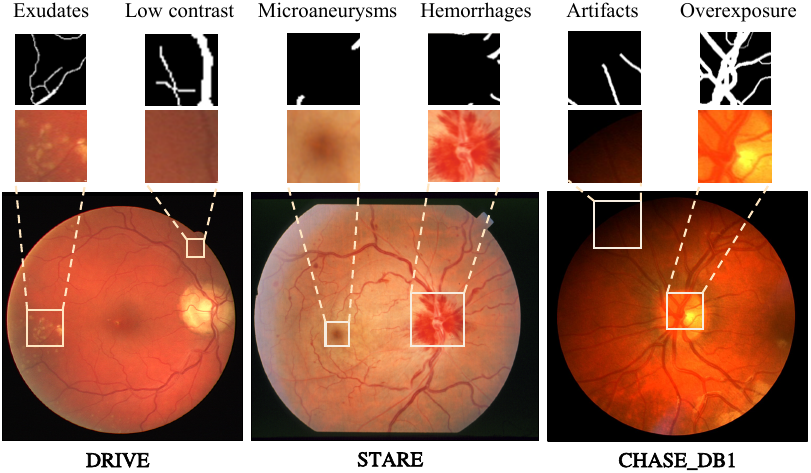}
\caption{Common challenges in retinal fundus image vessel segmentation.}
\label{fig:challenges}
\end{figure}
Although automatic methods address the limitations of manual segmentation, the task itself remains inherently complex. Achieving accurate and robust segmentation is complicated by several key challenges stemming from the complex nature of the fundus images themselves, as illustrated in Figure~\ref{fig:challenges}. First, poor image quality causes difficulties. Common problems include low contrast and overexposure. Both issues can obscure vessel details or result in missed fine vessels. Second, interference from pathologies creates obstacles. Lesions such as exudates, microaneurysms, and hemorrhages can blur vessel boundaries or mimic the appearance of blood vessels, increasing false positives and decreasing segmentation accuracy. Third, imaging artifacts add complications. The imaging process itself can cause distortions, which may compromise vascular structures and lead to fragmented segmentation results.

Deep learning techniques, particularly Convolutional Neural Networks (CNNs), have developed rapidly in recent years, demonstrating immense potential for addressing segmentation challenges. Among the multitude of deep learning models, U-Net \cite{2015U-Net} has become a foundational framework for retinal vessel segmentation due to its strong capabilities for local feature extraction. Building upon the U-Net architecture, several early variants emerged, including UNet++ \cite{2018UNet++}, Attention U-Net \cite{2018Attention}, R2U-Net \cite{2018R2U-Net}, and DUNet \cite{2019DUNet}. However, all listed models commonly struggle to capture long-range dependencies and offer insufficient nonlinear modeling capabilities. More recent studies continue to propose specialized convolutional variants. RVS-FDSC \cite{2024RVS-FDSC} introduces four-directional strip convolutions to simulate vessel paths. However, the reliance on fixed linear directions provides insufficient adaptability for tortuous vessels. TAOD-CFNet \cite{2025TAOD-CFNet} employs a Trumpet Attention Mechanism (TAM) to cross-fuse horizontal and vertical information from feature maps. Although the introduced TAM provides partial global information, the model's capacity for global modeling remains insufficient because the core extraction unit of TAOD-CFNet is still CNN-based. 

\begin{figure}[!htb]
    \centering
    \includegraphics[width=1.0\columnwidth]{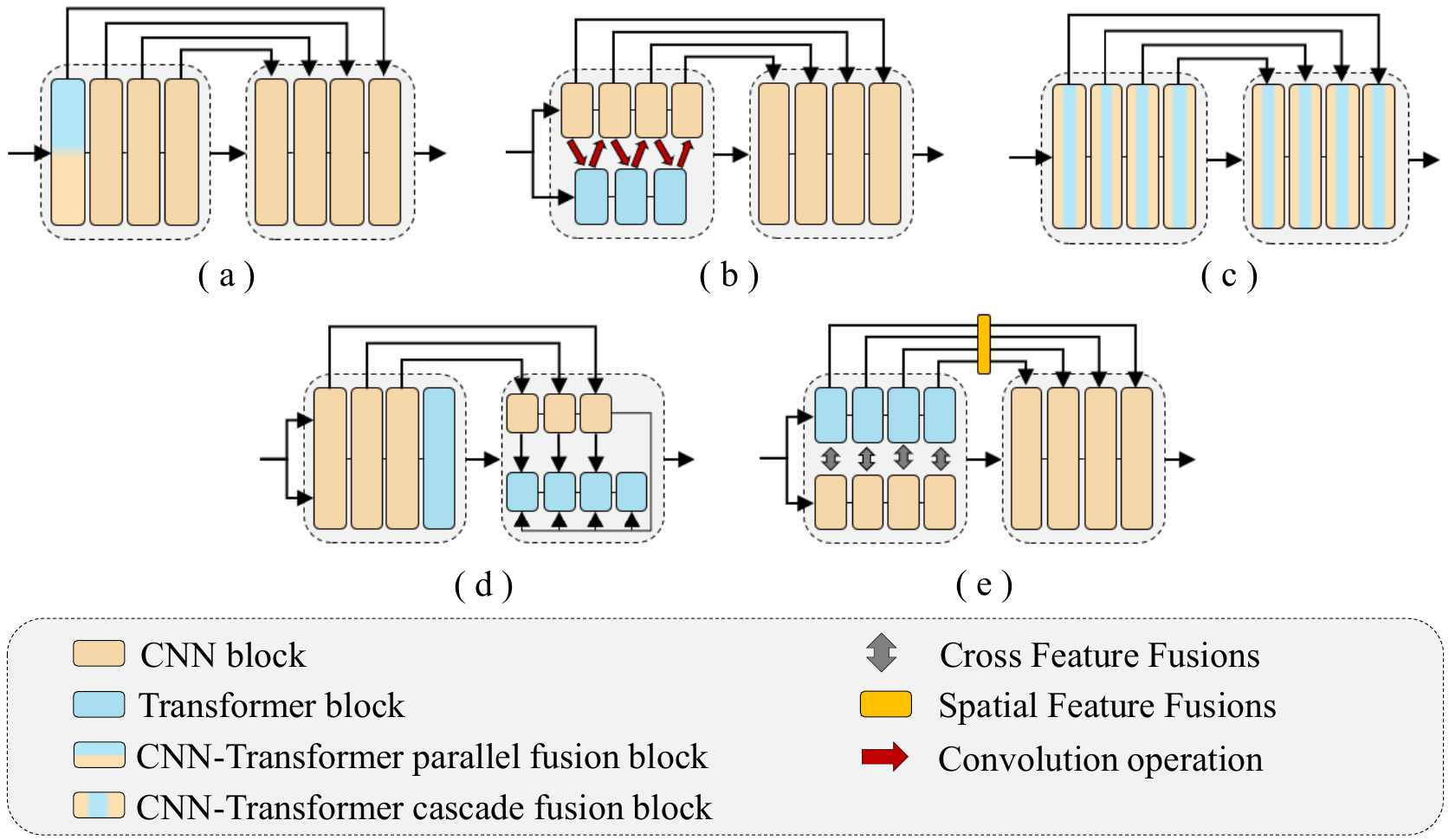}
    \caption{
        An illustration of typical hybrid CNN-Transformer architectures.
        \textbf{(a)} Shallow parallel fusion of CNN and Transformer branches.
        \textbf{(b)} Parallel dual-encoder with cross-level feature interaction.
        \textbf{(c)} Serial cascade fusion: CNN output feeds Transformer block.
        \textbf{(d)} Serial bottleneck fusion: Transformer embedded in U-Net bottleneck.
        \textbf{(e)} Parallel dual-encoder with same-level feature interaction.
        }
    \label{fig:hybrid_architectures}
\end{figure}

Transformers excel at capturing long-range dependencies, addressing a key limitation of conventional convolutional models. To fully leverage the distinct advantages of CNNs and Transformers, researchers began to develop hybrid CNN-Transformer architectures. Existing research has explored the fusion of the two architectures through diverse structural designs. For instance, SGAT-Net \cite{2023SGAT-Net} concentrates its innovation in the shallow layers of the encoder (as shown in Figure~\ref{fig:hybrid_architectures}(a)). The model employs a parallel structure to operate CNN and Transformer branches concurrently, fusing the local and global outputs within those shallow layers. A more thorough encoder modification is embodied in ARP-Net \cite{2023ARP-Net} (as shown in Figure~\ref{fig:hybrid_architectures}(b)), which features a parallel dual-branch structure that persists throughout the entire encoder. Information is exchanged between the two branches via a cross-level, interleaved Feature Interaction Unit (FIU). However, the ARP-Net design has a distinct computational sequence, potentially introducing limitations related to information latency. In contrast, HCTNet \cite{2023HCTNet} adopts a serial interleaved strategy (as shown in Figure~\ref{fig:hybrid_architectures}(c)), where the output of the CNN module within each level of the encoder and decoder is directly fed into a Transformer module. TransUNet \cite{2024TransUNet} cascades a CNN encoder into a Transformer bottleneck. The decoder features a complex parallel dual-branch structure to achieve high-precision segmentation (as shown in Figure~\ref{fig:hybrid_architectures}(d)). CFFormer \cite{2026CFFormer} presents another parallel dual-encoder model (as shown in Figure~\ref{fig:hybrid_architectures}(e)). The model employs a CFCA module to achieve same-level synchronous channel interaction, a design that contrasts sharply with the cross-level interaction of ARP-Net. Following the channel-level fusion, the XFF module is utilized to perform spatial feature fusion. Although progress has been made, the models discussed previously are still built upon conventional kernels and standard Multi-Layer Perceptron (MLP) layers, which struggle to capture the complex nonlinear relationships within fundus images.

Another significant deficiency of existing models is their neglect of interpretability, which heightens the risks associated with clinical decision-making. Recently, the Kolmogorov-Arnold Network (KAN) \cite{liu2024kAN} has offered a viable solution to the above problems, compensating for the deficiencies of traditional network structures with their superior nonlinear modeling capabilities and interpretability. For instance, U-KAN \cite{2025U-kan} embeds tokenized KAN blocks into the deep encoder and decoder, enabling the model to capture complex nonlinear relationships and improve interpretability. However, the model is not designed to capture long-range dependencies.

Beyond the challenges of nonlinear modeling and long-range dependencies, the fixed square receptive field of traditional convolution tends to introduce significant background noise when processing elongated and irregularly shaped targets, such as retinal vessels. In recent years, deformable convolution \cite{2017DCN1} and its variants \cite{2020DCN2, wang2023DCN3, 2024DCN4} have emerged to address the issue by learning two-dimensional offsets to locate target sampling points. The recently proposed Linear Deformable Convolution (LDConv) \cite{2023LDConv} builds upon prior work by breaking the constraints of the conventional square kernel, allowing the number of sampling points to be flexibly configured, which significantly reduces both computational and parameter overhead.

To address the aforementioned challenges, we propose a U-shaped model with a hybrid Transformer-CNN architecture named Adaptive Dual Branch Kolmogorov-Arnold UNet (DB-KAUNet). The DB-KAUNet introduces a Heterogeneous Dual-Branch Encoder (HDBE), which operates by alternating standard modules with Kolmogorov–Arnold modules. Specifically, the standard modules parallel a CNN block and a Transformer block, while the Kolmogorov–Arnold modules embed KAN-based layers within both pathways. The alternating design allows the HDBE to complementarily model features of retinal fundus images. The standard modules establish foundational features such as local vessel boundaries and the global vascular layout, while the Kolmogorov–Arnold modules precisely capture the complex and tortuous morphology of fine retinal vessels. Concurrently, several specialized components are embedded within the encoder to optimize the integration of retinal fundus image features. A Cross-Branch Channel Interaction (CCI) module is introduced to facilitate efficient channel-level feature interaction between the parallel pathways. An attention-based Spatial Feature Enhancement (SFE) module is also included to enhance spatial features and fuse the final outputs from both encoder branches. We further advance the SFE module by replacing standard convolutions with LDConv, creating the Spatial Feature Enhancement with Geometrically Adaptive Fusion (SFE-GAF) module. The SFE-GAF module utilizes an X-shaped sampling pattern that is specifically designed to adaptively focus computational resources on the actual morphology of the vessels, enhancing feature accuracy while using fewer parameters.

In summary, the main contributions of this research are as follows:
\begin{itemize}
    \item We propose DB-KAUNet, a novel hybrid architecture that integrates the powerful nonlinear modeling capabilities of KAN into a parallel CNN-Transformer dual-encoder framework for retinal vessel segmentation.
    \item We present an HDBE employing an innovative alternating design of standard and Kolmogorov-Arnold modules. The design allows the model to establish the foundational local features and global vascular topology, as well as to precisely capture the complex nonlinear morphology of fine vessels.
    \item We introduce a CCI module for channel-level interaction and SFE/SFE-GAF modules for spatial-level fusion. These modules resolve the inherent feature disparities between the parallel CNN and Transformer encoders, producing a unified and robust feature representation that is critical for segmenting the complex and continuous morphology of retinal vessels.
    \item Our DB-KAUNet achieves state-of-the-art performance on the DRIVE, STARE, and CHASE\_DB1 datasets, demonstrating superior generalization and robust capabilities, particularly in the challenging task of segmenting fine micro-vessels.
\end{itemize}

The remainder of this paper is organized as follows. Section~\ref{sec:methodology} details our proposed method. The experimental setup is given in Section~\ref{sec:experiments}.
Section~\ref{sec:results_and_discussion} presents the experimental results and discussion. Finally, Section~\ref{sec:Conclusion} concludes the paper.

\section{Methodology}
\label{sec:methodology}

\begin{figure*}[!htb]
    \centering
    \includegraphics[width=\textwidth]{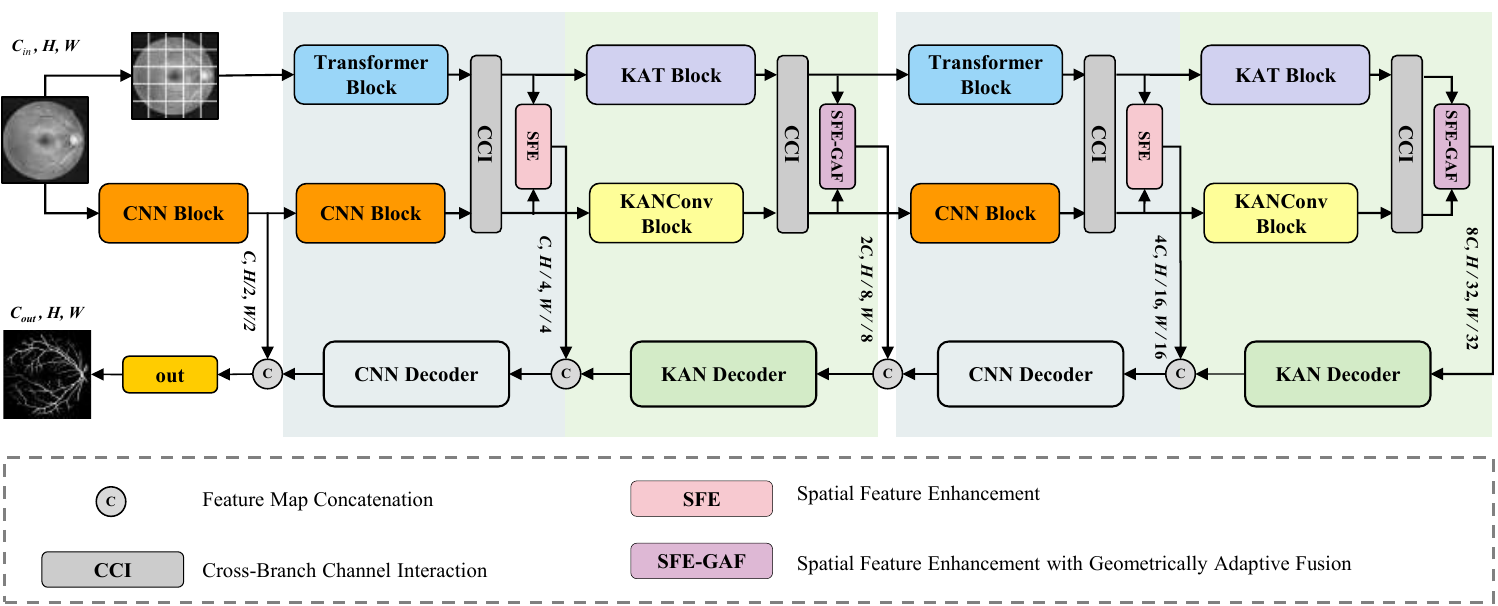}
    \caption{The overall architecture of the proposed DB-KAUNet.}
    \label{fig:architecture}
\end{figure*}

\subsection{Preliminary:KAN}
\label{subsec:Kolmogorov–Arnold_Networks}
The Kolmogorov-Arnold representation theorem states that any multivariate continuous function can be represented as a finite superposition of univariate functions and the operation of addition. Specifically, a multivariate continuous function $f(x_1, \dots, x_n)$ can be represented as:
\begin{equation}
f(x_1, \dots, x_n) = \sum_{q=0}^{2n+1} \Phi_q \left( \sum_{p=1}^{n} \phi_{q,p}(x_p) \right)
\end{equation}
where $\Phi_q$ and $\phi_{q,p}$ are continuous, single-variable functions.

Inspired by the Kolmogorov-Arnold representation theorem, Liu et al. \cite{liu2024kAN} proposed the KAN. In contrast to the MLP which uses fixed activation functions on nodes, KAN features learnable univariate activation functions on their edges. This shifts the learning paradigm from learning the weights of linear transformations to learning the activation functions on the edges of the network.

The activation of the $j$-th neuron in the $(l+1)$-th layer is given by:
\begin{equation}
x_{l+1,j} = \sum_{i=1}^{n_{l}} \phi_{l,j,i}(x_{l,i})
\end{equation}
where $x_{l,i}$ is the activation of the $i$-th neuron in layer $l$, $n_l$ is the number of nodes in layer $l$, and $\phi_{l,j,i}$ is the learnable activation function on the edge connecting the $i$-th neuron of layer $l$ to the $j$-th neuron of layer $l+1$.

The matrix form is given by:
\begin{equation}
\label{matrix}
\mathbf{x}_{l+1} = \underbrace{
\begin{pmatrix}
\phi_{l,1,1}(\cdot) & \phi_{l,1,2}(\cdot) & \cdots & \phi_{l,1,n_l}(\cdot) \\
\phi_{l,2,1}(\cdot) & \phi_{l,2,2}(\cdot) & \cdots & \phi_{l,2,n_l}(\cdot) \\
\vdots & \vdots & \cdots & \vdots \\
\phi_{l,n_{l+1},1}(\cdot) & \phi_{l,n_{l+1},2}(\cdot) & \cdots & \phi_{l,n_{l+1},n_l}(\cdot)
\end{pmatrix}
}_{\Phi_l} \mathbf{x}_l
\end{equation}
where $\mathbf{x}_l \in \mathbb{R}^{n_l}$ and $\mathbf{x}_{l+1} \in \mathbb{R}^{n_{l+1}}$ are the activation vectors of layer $l$ and $l+1$ respectively, and $\Phi_l$ (with $\Phi_{l} = \{\phi_{l,j,i}\}$) is the function matrix of the $l$-th KAN layer.

The composition of KAN layers forms a universal KAN. Given an input vector, the output of the KAN is:
\begin{equation}
KAN(\mathbf{x}_{0}) = (\Phi_{L-1} \circ \Phi_{L-2} \circ \cdots \circ \Phi_1 \circ \Phi_0) (\mathbf{x}_{0})
\end{equation}
where the $\circ$ denotes the composition of functions.

\subsection{Overall Architecture}
\label{subsec:Overall Architecture}

The architecture of our proposed DB-KAUNet model is shown in Figure~\ref{fig:architecture}. The model adopts a U-shaped architecture and introduces a novel HDBE. The HDBE employs two parallel encoders: a CNN encoder and a Transformer encoder. The CNN encoder is categorized into standard CNN blocks and our KANConv blocks. Similarly, the Transformer encoder is subdivided into the standard Transformer and the Kolmogorov-Arnold Transformer (KAT) proposed by Yang and Wang \cite{2024kat}. Within the HDBE, the standard modules and the Kolmogorov–Arnold modules are interleaved. The HDBE also integrates a CCI module for feature interaction between the parallel pathways and an SFE module to process features for the skip connections. Building upon the SFE module, an SFE-GAF module is developed, which replaces the standard convolution with LDConv. Following a similar interleaving strategy, the SFE module is embedded within the standard modules, and the SFE-GAF module is embedded within the Kolmogorov–Arnold modules.

\subsection{Heterogeneous Dual-Branch Encoder}
\label{subsec:Heterogeneous Dual-Branch Encoder}

\begin{figure*}[!htb] 
    \centering 
    \includegraphics[width=\textwidth]{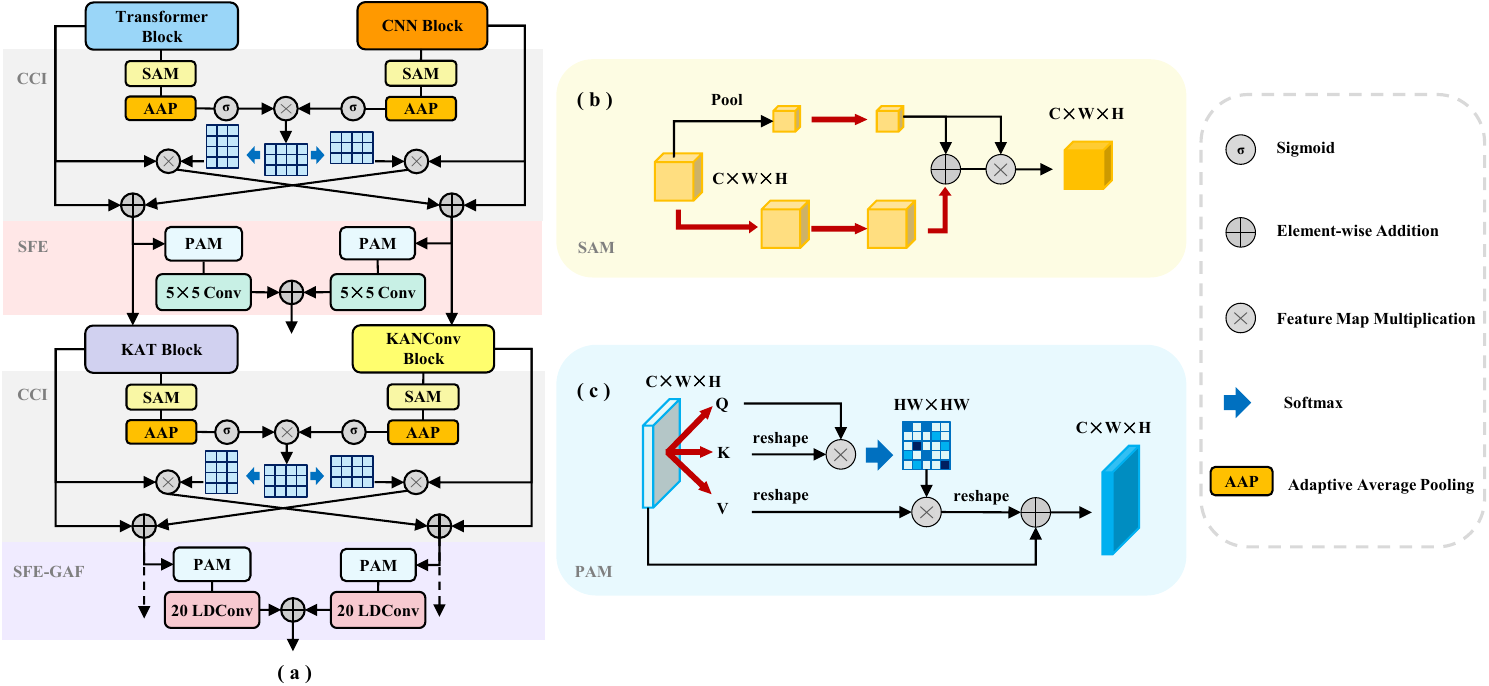} 
    \caption{Illustration of the HDBE and its components. (a) The structure of the HDBE. (b) The Squeeze-and-Attention Module. (c) The Position Attention Module.}
    \label{fig:encoder} 
\end{figure*}

As illustrated in Figure~\ref{fig:encoder}, the proposed HDBE is composed of a five-layer architecture. In the first layer, a 7x7 convolution is employed as a standard CNN block for feature extraction. The second and fourth layers are designated as standard modules, while the third and fifth layers are designated as Kolmogorov-Arnold modules.

\subsubsection{Standard Modules}
In our standard module, a residual block from the Residual Network (ResNet) \cite{2016ResNet} is adopted as the CNN block, and the Vision Transformer (ViT) \cite{2021Vit} is employed as the Transformer block. This approach is designed to utilize the CNN for local feature extraction and the Transformer for global feature modeling. Leveraging the low computational cost of fixed activation functions like ReLU, the module efficiently extracts foundational, general-purpose features.
\subsubsection{Kolmogorov–Arnold Modules}

\begin{figure}[!htb]
    \centering
    \includegraphics[width=1.0\columnwidth]{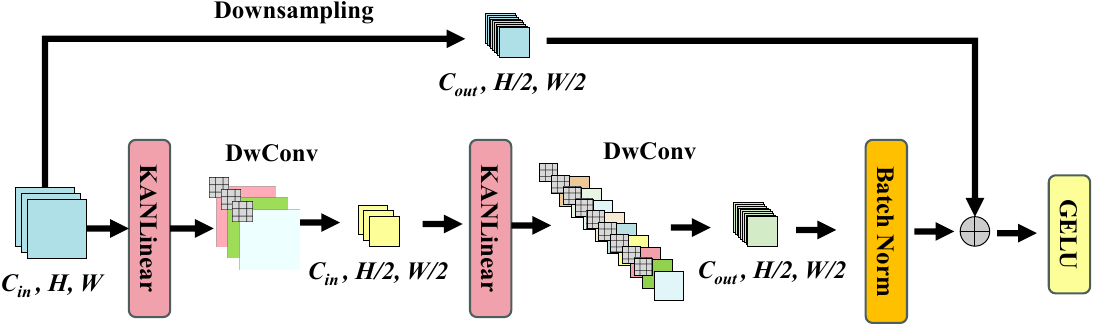}
    \caption{Detailed structure of the proposed KANConv block.}
    \label{fig:kan_block}
\end{figure}

As illustrated in Figure~\ref{fig:kan_block}, the proposed KANConv block processes an input feature map through two parallel pathways: a main transformation path and a shortcut path. The main transformation path consists of sequential KANLinear layers and Depthwise Convolutions \cite{2022DWConv}. The outputs of both pathways are summed element-wise and passed through a GELU activation function to produce the final output.

Formally, let the input feature map be denoted by $X \in \mathbb{R}^{C \times H \times W}$. The output of the KANConv block, $Y \in \mathbb{R}^{2C \times H/2 \times W/2}$, is given by:
\begin{equation} \label{eq:main_block}
Y = GELU(H(X) + D(X))
\end{equation}
where $H(X)$ represents the sequence of nonlinear transformations in the main path. $D(X)$ represents the transformation on the shortcut path, which performs downsampling and channel matching to align with the main path.

The transformation $H(X)$ along the main path is a composite function, formally expressed as:
\begin{equation} \label{eq:main_branch_comp}
H(X) = (BN \circ DwConv \circ \Phi \circ DwConv \circ \Phi)(X)
\end{equation}
where $\Phi$ is the KANLinear layer, which is designed to model complex nonlinear relationships using learnable activation functions.

The learnable activation function $\phi(x)$ on each edge is formulated as a weighted sum of a fixed base function and a linear combination of B-spline basis functions:
\begin{equation} \label{eq:kan_layer}
\phi(x) = \beta  SiLU(x) + s  \sum_{k=1}^{N} c  B_k(x)
\end{equation}
where $\beta$, $s$, and $c$ are learnable parameters, $SiLU(\cdot)$ \cite{2018SiLU} is the activation function, the $B_k$ are B-spline basis functions, and $N$ is the number of B-spline basis functions.

\begin{figure}[!htb]
\centering
\includegraphics[width=1.0\columnwidth]{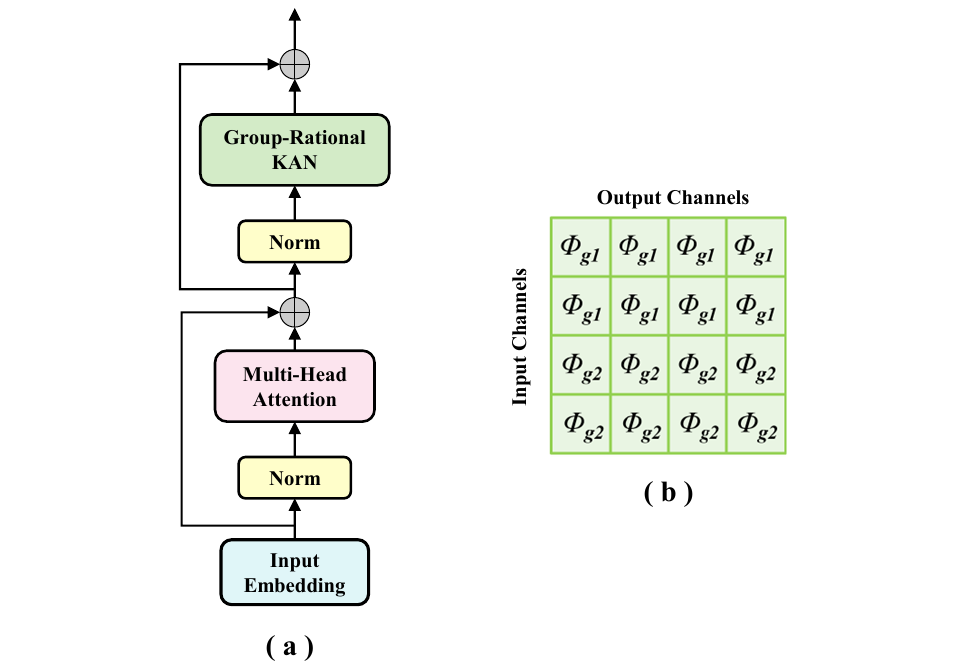}
\caption{Illustration of the KAT block. (a) The structure of the KAT block. (b) The GR-KAN layer.}
\label{fig:kat_block_detail}
\end{figure}

Running in parallel to the KANConv block is a KAT block \cite{2024kat}, which is formed by strategically replacing the feed-forward MLP layer in a ViT with the Group-Rational KAN (GR-KAN) layer. The design of the GR-KAN layer, as shown in Figure~\ref{fig:kat_block_detail}, is suited to the specific requirements of the Transformer architecture. As computationally intensive, large-scale models, Transformers demand both powerful global fitting capabilities and high computational efficiency. To satisfy the global fitting requirement, the use of rational functions is highly effective. Rational functions are composed of polynomials, offering robust fitting capabilities, whereas B-spline functions require recursive computation, making them less suitable for integration into the Transformer architecture. To satisfy the efficiency requirement, the strategy of sharing learnable parameters among channel groups is employed.

Specifically, the learnable function $\phi(x)$ on each edge is parameterized as a rational function, which consists of an m-th order polynomial $P(x)$ and an n-th order polynomial $Q(x)$. To avoid the numerical instability caused by poles, which occur when the denominator $Q(x) \to 0$ and $\phi(x) \to \pm\infty$, a Safe Pad\'e Activation Unit (PAU) \cite{2020Pad} is further employed as the implementation method.
\begin{equation} \label{eq:kat}
\phi(x) = \gamma F(x) = \gamma \frac{P(x)}{Q(x)} = \gamma \frac{a_0 + a_1x + \dots + a_m x^m}{1 + |b_1x + \dots + b_n x^n|}
\end{equation}
where $a_m$, $b_n$ are the coefficients of the rational function and $\gamma$ is a scaling factor, all of which are learnable parameters. 

\subsection{Cross-Branch Channel Interaction Module}
\label{subsec:Cross-Branch Channel Interaction}

To effectively integrate the local and global features within retinal fundus images, the CCI module is proposed. The module establishes a bidirectional information flow. On one hand, rich local structural information is transferred from the CNN branch to the Transformer branch. On the other hand, global contextual semantics are passed from the Transformer branch to the CNN branch.

To enable effective interaction, the CCI module must first refine the distinct feature representations arriving from the parallel encoder pathways. Therefore, A Squeeze-and-Attention Module (SAM) \cite{2020Squeeze-and-attention} is employed to process the multi-channel local and global feature maps. Since SAM processing does not change the feature map dimensions, the resulting features remain high-dimensional. Constructing a correlation matrix directly from the high-dimensional features would cause significant computational overhead and lower attention efficiency. To overcome the computational bottleneck, Adaptive Average Pooling (AAP) is employed to efficiently compress the feature maps into vectors that represent the characteristics of each channel. The resulting vectors are passed through a Sigmoid function to generate the final channel attention vectors, as formulated below:
\begin{equation}
\begin{split}
L_{attn} &= \sigma(AAP(SAM(L))) \\
G_{attn} &= \sigma(AAP(SAM(G)))
\end{split}
\end{equation}
where $L \in \mathbb{R}^{C_{c} \times H \times W}$ is the multi-channel local feature map from the CNN branch, and $G \in \mathbb{R}^{C_{t} \times H \times W}$ is the global feature map from the Transformer branch. $C_{c}$ and $C_{t}$ are the channel counts, $H$ and $W$ are the spatial dimensions, and $\sigma(\cdot)$ denotes the Sigmoid function.

Next, the cross-feature channel correlation matrix $R \in \mathbb{R}^{C_c \times C_t}$, which captures the correlations between the multi-channel features, is computed via matrix multiplication. As illustrated by the blue grids in Figure~\ref{fig:encoder}(a), the formula is as follows:
\begin{equation}
R = L_{attn} \otimes G_{attn}^T
\end{equation}

The correlation matrix $R$ is employed as a transformation matrix. Its purpose is to perform projective transformations on the feature maps $L$ and $G$, respectively. These transformations permit subsequent feature fusion. In this process, $L$ is projected into a new tensor with the same dimensions as $G$. Concurrently, $G$ is projected into a new tensor with the same dimensions as $L$. These projection operations are formulated as follows:
\begin{equation}
\begin{split}
L_{\to G} &= L \otimes_1 Softmax(R) \\
G_{\to L} &= G \otimes_1 Softmax(R^T)
\end{split}
\end{equation}
where $Softmax(\cdot)$ is the activation function used to normalize the channel correlation matrix $R$. The term $\otimes_1$ denotes the 1-mode tensor product \cite{2009mode}. It is worth noting that the normalization dimension of Softmax is not fixed but is adaptively determined by the number of input feature channels (as shown by the blue arrows in Figure~\ref{fig:encoder}). For example, when the input is the feature map $L$ with $C_c$ channels, Softmax is applied along the $C_c$ dimension of the correlation matrix, and vice versa. The subsequent 1-mode tensor product operation is essentially a matrix multiplication between the first dimension of the 3D tensor ($L$ or $G$) and the 2D correlation matrix $R$. This operation ultimately generates new feature maps $L_{\to G} \in \mathbb{R}^{C_t \times H \times W}$ and $G_{\to L} \in \mathbb{R}^{C_c \times H \times W}$.

Bidirectional feature fusion is achieved through an additive operation. On one hand, the projected features $G_{\to L}$, which carry global context, are added to $L$. This addition enhances the global perception capability of the CNN branch. On the other hand, the projected features $L_{\to G}$, which are rich in local details, are added to $G$. This process supplements the local information of the Transformer branch. The specific fusion process is as follows:
\begin{equation}
\begin{split}
L_{fuse} &= G_{\to L} + L \\
G_{fuse} &= L_{\to G} + G
\end{split}
\end{equation}

By continuously performing this interaction within the encoder, the module not only mitigates the discrepancies in channel features between the two pathways but also significantly enhances model robustness. Consequently, the model can accurately differentiate true vascular structures from background noise or pathological lesions, even when processing low-quality images with issues such as blurriness and artifacts.

\subsection{Spatial Feature Enhancement Module}
\label{subsec:Spatial Feature Enhancement module}

To reduce the differences in the spatial representations of the retinal fundus images between the two encoder branches, the SFE module is proposed. The module begins by processing the outputs from the CCI module using the Position Attention Module (PAM) \cite{2019PAM} to model the long-range spatial dependencies inherent in the continuous vessel pathways. Next, a $5 \times 5$ convolution is applied to each pathway to unify the channel dimensions. The two results are then fused via element-wise summation to generate the skip connection input $X_{skip}$, as formulated below:
\begin{equation}
X_{skip} = Conv_{5\times5}(PAM(L_{fuse})) + Conv_{5\times5}(PAM(G_{fuse})) 
\end{equation}

This design elegantly combines the local details captured by the convolution with the global contextual information extracted by the PAM. The resulting spatially refined representation effectively captures both the precise boundaries and the continuous pathways of retinal vessels.

\begin{figure}[!htb]
    \centering
    \includegraphics[width=1.0\columnwidth]{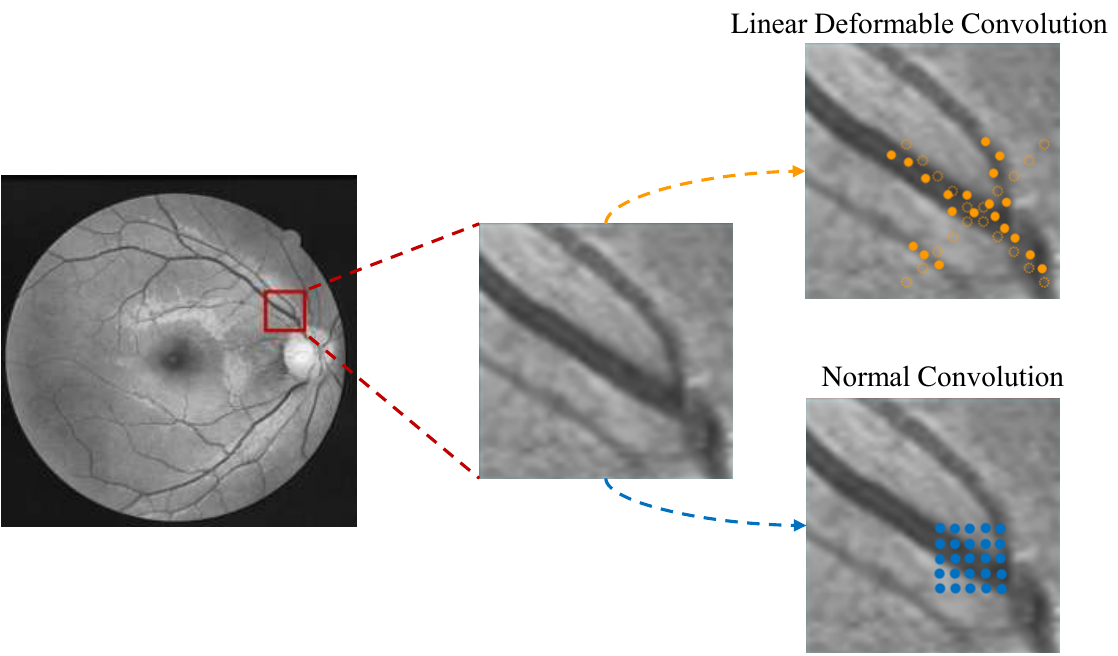}
    \caption{Comparison of Sampling Patterns between Linear Deformable Convolution and Normal Convolution.}
    \label{fig:LDConv}
\end{figure}
To better adapt to the complex morphology of slender and tortuous retinal vessels, we modified the original SFE module. The core modification is replacing the fixed $5 \times 5$ convolution within the SFE module with the more flexible LDConv \cite{2023LDConv}. An X-shaped initial sampling pattern is designed for the LDConv. Illustrated in Figure~\ref{fig:LDConv}, the pattern consists of four slender branches that total 20 sampling points and are shaped to better match the vascular structures. By learning offsets, the sampling branches can dynamically bend, thereby achieving geometric adaptation to vessel trajectories. The new module with geometric adaptive fusion capabilities is named SFE-GAF. It is used to replace the original SFE module within the encoder's Kolmogorov-Arnold modules. Ultimately, the module's capability to capture complex vascular morphological features is significantly enhanced.

Standard convolution typically uses the center of the sampling grid as the origin $(0,0)$, whereas LDConv defines the top-left position as the sampling origin. Based on this coordinate system, the initial set of sampling points $S$ that forms the X-shaped pattern is represented as:
\begin{equation}
S = \{ (0, 0), (1, 1), \dots, (9, 9)\} \cup \{(0, 9), (1, 8), \dots, (9, 0) \}
\end{equation}

In LDConv, this initial set of sampling points is dynamically adjusted by learning offsets to obtain optimal sampling locations. Therefore, the computation at any position on the output feature map can be formulated as:
\begin{equation}
Y(p_0) = \sum_{p_n \in S} w_n  X(p_0 + p_n + \Delta p_n)
\end{equation}
where $Y(p_0)$ is the output value at position $p_0$, $X$ is the input feature map, $p_n$ denotes each sampling point in $S$, $w_n$ is the learnable weight for $p_n$, and $\Delta p_n$ is the learned offset for that point. Since the learned offsets $\Delta p_n$ result in floating-point sampling coordinates, bilinear interpolation is employed to accurately sample the corresponding feature values.

\subsection{Decoder}
\label{subsec:Decoder}

\begin{figure}[!htb]
    \centering
    \includegraphics[width=1.0\columnwidth]{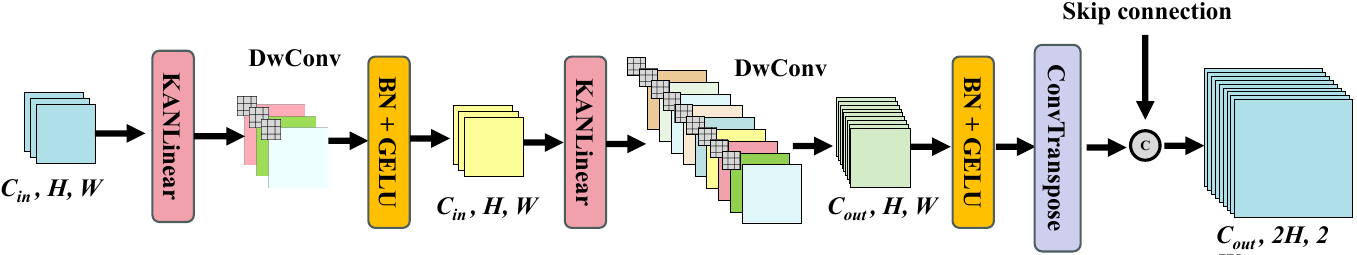}
    \caption{The structure of the KAN Decoder.}
    \label{fig:upsample}
\end{figure}
Echoing the encoder, the decoder also adopts an interleaved design. This design consists of two different types of decoder blocks arranged alternately: the CNN Decoder and the KAN Decoder. The CNN Decoder is structurally similar to a standard U-Net decoder block, employing a transposed convolution followed by a dual standard CNN block. In contrast, the KAN Decoder replaces standard CNNs with Depthwise Convolutions to reduce the number of parameters. A KANLinear layer is also embedded within the KAN Decoder before each convolution to enhance nonlinear capabilities. The structure of the KAN Decoder is detailed in Figure~\ref{fig:upsample}.

Feature fusion in the decoder is handled by concatenation. The input to each decoder layer, excluding the deepest one, is formed by two elements: the output of the preceding layer and the skip connection from the corresponding encoder layer. An interleaved skip-connection strategy is employed for all layers beyond the first. The purpose of this strategy is to achieve complementarity between basic and complex nonlinear features. Specifically, the output of a KAN Decoder is concatenated with the output from a standard module in the encoder. Conversely, the output of a CNN Decoder is concatenated with the output from a Kolmogorov-Arnold module in the encoder.

\subsection{Loss Function}
\label{subsec:loss_function}
The segmentation of retinal vessels is a classic problem characterized by class imbalance, where background pixels vastly outnumber vessel pixels. To address this, a composite loss function is employed that combines the strengths of pixel-wise Cross-Entropy (CE) and the Dice coefficient. The CE loss, $\mathcal{L}_{CE}$, evaluates the per-pixel classification error, ensuring that each pixel is correctly labeled. The Dice loss \cite{2016Dice}, $\mathcal{L}_{Dice}$, derived from the Dice coefficient, directly optimizes the overlap between the prediction and the ground truth, making it robust to class imbalance. Our total loss is a weighted sum of these two components:
\begin{equation}
    \mathcal{L}_{total} = \alpha \mathcal{L}_{CE} + (1-\alpha) \mathcal{L}_{Dice}
\end{equation}
where $\alpha$ is a hyperparameter balancing the two terms, which we empirically set to 0.5 to give equal importance to both pixel-level accuracy and region-based overlap.

\section{Experiments}
\label{sec:experiments}
\subsection{Datasets}
\label{subsec1:datasets}

To evaluate the proposed model, we selected three publicly available retinal vessel segmentation datasets (DRIVE, STARE, and CHASE\_DB1) for testing. A brief introduction to these datasets is provided below.

(1) \textbf{DRIVE}: The Digital Retinal Images for Vessel Extraction (DRIVE) dataset is a publicly available collection for retinal vessel segmentation derived from a diabetic retinopathy screening program in the Netherlands \cite{2004DRIVE}. It contains 40 color retinal fundus images with a resolution of 565 $\times$ 584 pixels, seven of which show signs of diabetic retinopathy.

(2) \textbf{STARE}: The Structured Analysis of the Retina (STARE) dataset was first cited and made publicly available by Hoover et al. \cite{Hoover2000STARE}. The dataset contains 20 color retinal fundus images, each with a resolution of 700 $\times$ 605 pixels. Half of the images show healthy retinas, and the other half show different types of retinal diseases.

(3) \textbf{CHASE\_DB1}: The Child Heart and Health Study in England (CHASE\_DB1) dataset was derived from a health study of 200 elementary schools in the U.K. \cite{2009CHASE}. It contains 28 color fundus photographs with a resolution of 999 $\times$ 960 pixels. For each photograph, the corresponding vessel segmentation map has been meticulously labeled by experts to facilitate model training and evaluation.

\subsection{Implementation Details}
\label{subsec3:implementation_details}
In terms of standardization, we extracted the green channel from each RGB fundus image, which provided the highest contrast between the foreground and background. We then applied Contrast Limited Adaptive Histogram Equalization (CLAHE) to address variations in illumination and contrast. Subsequently, the processed grayscale images were normalized to the range of [0, 1]. Finally, gamma correction was applied to further standardize the illumination. All evaluations were conducted strictly within the provided Field of View (FOV) masks to exclude interference from non-retinal regions.

For data augmentation, we employed a strategy of random rotations, horizontal and vertical flips, and color jittering to enhance the model's generalization ability and prevent overfitting.

For image patch extraction, we trained the model on 150,000 patches of size 64 $\times$ 64 pixels randomly sampled from the training images in each dataset. At test time, we employed a sliding window strategy with a stride of 8 pixels to scan the entire image. The final prediction was generated by aggregating the predictions from all extracted patches.

All experiments were conducted on a single NVIDIA RTX 5090 GPU using the PyTorch framework. We set the number of training epochs to 50 and the batch size to 64. The AdamW optimizer was employed with a weight decay of $1 \times 10^{-5}$ and an initial learning rate of $5 \times 10^{-4}$. We utilized a cosine annealing schedule to dynamically adjust the learning rate. To ensure training stability, gradient clipping was applied after backpropagation with a maximum norm set to 5.0. We designated 10\% of the training data as the validation set and implemented an early stopping mechanism with a patience of 10 epochs to prevent overfitting. The model's performance was assessed using several key metrics, including Area Under the Curve (AUC), Sensitivity (SE), Specificity (SP), Accuracy (ACC), and the F1 score (F1). Finally, the model checkpoint that achieved the highest F1 on the validation set was saved and used for the final evaluation on the test set.

\section{Results And Discussion}
\label{sec:results_and_discussion}

\subsection{Ablation Study}
\label{subsec:ablation_study}

\begin{figure*}[!htbp]
    \centering
    \includegraphics[width=\textwidth]{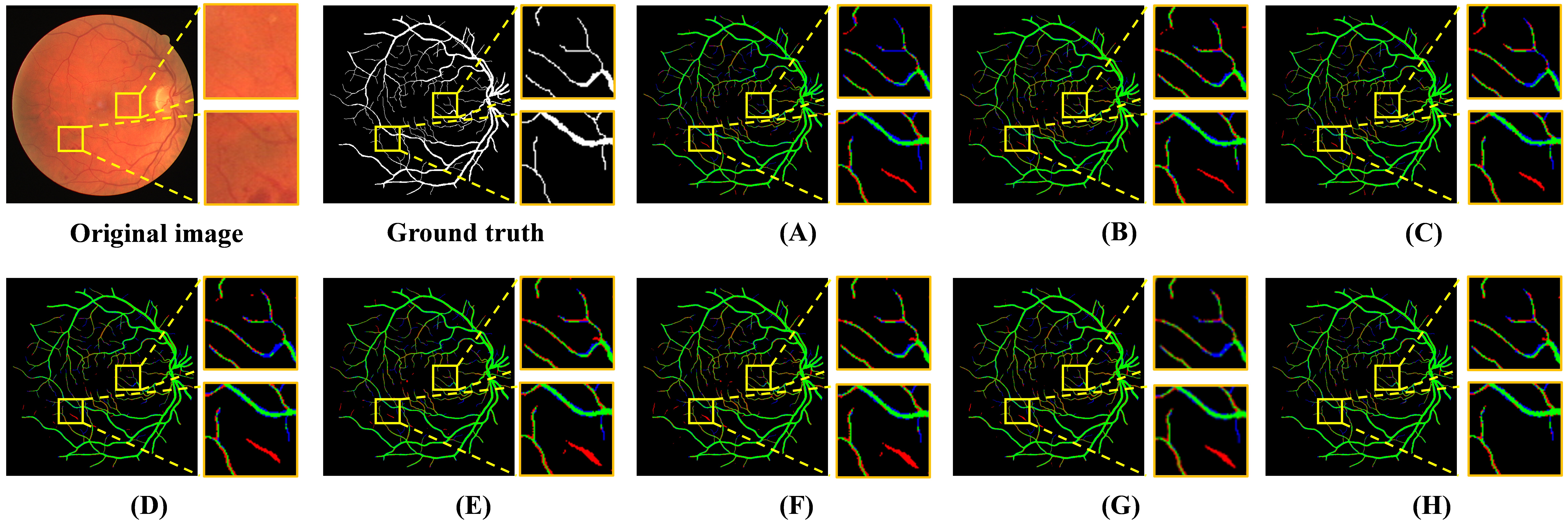} 
    \caption{Ablation Study on the DRIVE Dataset. The figure presents the segmentation results of different model configurations described in Table~\ref{tab:ablation_study}. In these results, green pixels represent true positives (TP), red represents false positives (FP), and blue represents false negatives (FN). This visual comparison highlights the progressive improvements in segmentation as our proposed modules are integrated.}
    \label{fig:ablation_study_comparison}
\end{figure*}

\begin{table}[!htbp]
    \centering
    \caption{Ablation Study of DB-KAUNet Components on the DRIVE Dataset. Best results are in \textcolor{red}{\textbf{red}}, second best in \textcolor{blue}{blue}.}
    \label{tab:ablation_study}
    \resizebox{\columnwidth}{!}{
    \begin{tabular}{lcccc}
        \toprule
        \textbf{Model Configuration} & \textbf{F1} & \textbf{SE} & \textbf{SP} & \textbf{ACC} \\
        \midrule
        (A) U-Net & 0.8108 & 0.7776 & \best{0.9867} & 0.9681 \\
        (B) CNN + Transformer + Decoder & 0.8688 & 0.8711 & 0.9808 & 0.9671 \\
        (C) CNN + Transformer + KAN Decoder & 0.8711 & 0.8718 & 0.9815 & 0.9677 \\
        (D) HDBE + KAN Decoder & 0.8756 & 0.8791 & 0.9816 & 0.9687 \\
        (E) HDBE + CCI + KAN Decoder & 0.8759 & 0.8811 & 0.9815 & 0.9692 \\
        (F) HDBE + SFE + KAN Decoder & 0.8721 & 0.8735 & 0.9814 & 0.9679 \\
        (G) HDBE + CCI + SFE + KAN Decoder & \secondbest{0.8812} & \secondbest{0.8826} & 0.9827 & \secondbest{0.9701} \\
        (H) HDBE + CCI + SFE + GAF + KAN Decoder & \best{0.8964} & \best{0.8985} & \secondbest{0.9848} & \best{0.9739} \\
        \bottomrule
    \end{tabular}}
\end{table}

We conducted a comprehensive ablation study on the DRIVE dataset to systematically validate the contributions of each component within the proposed DB-KAUNet. The quantitative results are presented in Table~\ref{tab:ablation_study}, and visual comparisons are shown in Figure~\ref{fig:ablation_study_comparison}.

We first established a baseline model using a U-Net with a standard CNN encoder-decoder architecture. Subsequently, we designed and tested a dual-encoder architecture, Model (B), that paralleled a residual block from ResNet and a ViT at each encoder layer, attempting to fuse their features through simple concatenation. The experimental results in Table~\ref{tab:ablation_study} indicated that Model (B) achieved a significant improvement in the F1 over the baseline Model (A), confirming the benefit of capturing long-range dependencies. However, while CNNs exhibit strong noise robustness via local receptive fields, the core self-attention mechanism of the Transformer is highly susceptible to misinterpreting noise and artifacts in low-quality images as salient features. This susceptibility leads to an increase in false positives, a fact confirmed by the significantly lower SP value of Model (B) compared to Model (A). This outcome demonstrates that this simple fusion approach is a suboptimal trade-off, as the naive integration not only impairs the ability to identify negative samples but also fails to bridge the significant domain gap between the two feature types.

To address the limitations of the simple dual-encoder design, we evaluated the impact of our KAN-based modules. We tested Model (C) which integrates the KAN Decoder and Model (D) which utilizes both the HDBE and the KAN Decoder. To clearly see the effect of the KAN components, both Model (C) and Model (D) used the exact same simple concatenation fusion as Model (B). Compared to Model (B), Model (C) achieved slight gains, increasing the F1 by 0.23\% and SE by 0.07\%. Model (D) achieved significant gains, increasing the F1 by 0.68\% and SE by 0.80\%. We attribute this marked increase in SE to the superior nonlinear modeling capability of both the Kolmogorov–Arnold modules and the KAN Decoder, which allows the model to precisely capture fine micro-vessel details and effectively mitigate class imbalance.

Building on Model (D), we then evaluated our two synergistic fusion modules: the CCI and the SFE. Our ablation study validates this dual-component design. Model (E) (CCI only) improved performance over Model (D), demonstrating the effectiveness of channel interaction. Conversely, Model (F) (SFE only) caused a slight performance degradation compared to Model (D). We attribute this degradation to the SFE receiving redundant channel features that are unfiltered by the CCI. These features were then repeatedly amplified by the convolutional operations within the SFE, ultimately leading the output to deviate from the task objective. Model (G), which integrated both CCI and SFE, achieved a notable performance increase, thereby confirming the synergistic relationship between CCI and SFE. It significantly outperformed Model (D) with improvements in F1, SE, and ACC of 0.56\%, 0.35\%, and 0.14\%, respectively. This outcome confirms that mitigating feature disparities requires addressing both channel and spatial inconsistencies simultaneously.

To better adapt to the unique morphological and topological structures of retinal vessels, we further propose the SFE-GAF module. This module is realized by replacing the standard convolution within the SFE module with LDConv. This improvement enables the sampling points of the convolutional kernel to dynamically conform to the paths and shapes of the vessels, thereby achieving more precise extraction of geometric features. Our final Model (H), built upon the architecture of Model (G), achieves peak performance by interleaving the standard SFE and the SFE-GAF modules. Compared to the initial baseline Model (A), Model (H) achieves significant improvements of 8.56\% in F1, 12.09\% in SE, and 0.58\% in ACC.

\subsection{Comparison with State-of-the-Art Methods}
\label{subsec:Comparison_with_state-of-the-art_methods}

\begin{figure*}[!htb] 
    \centering
    \includegraphics[width=\textwidth]{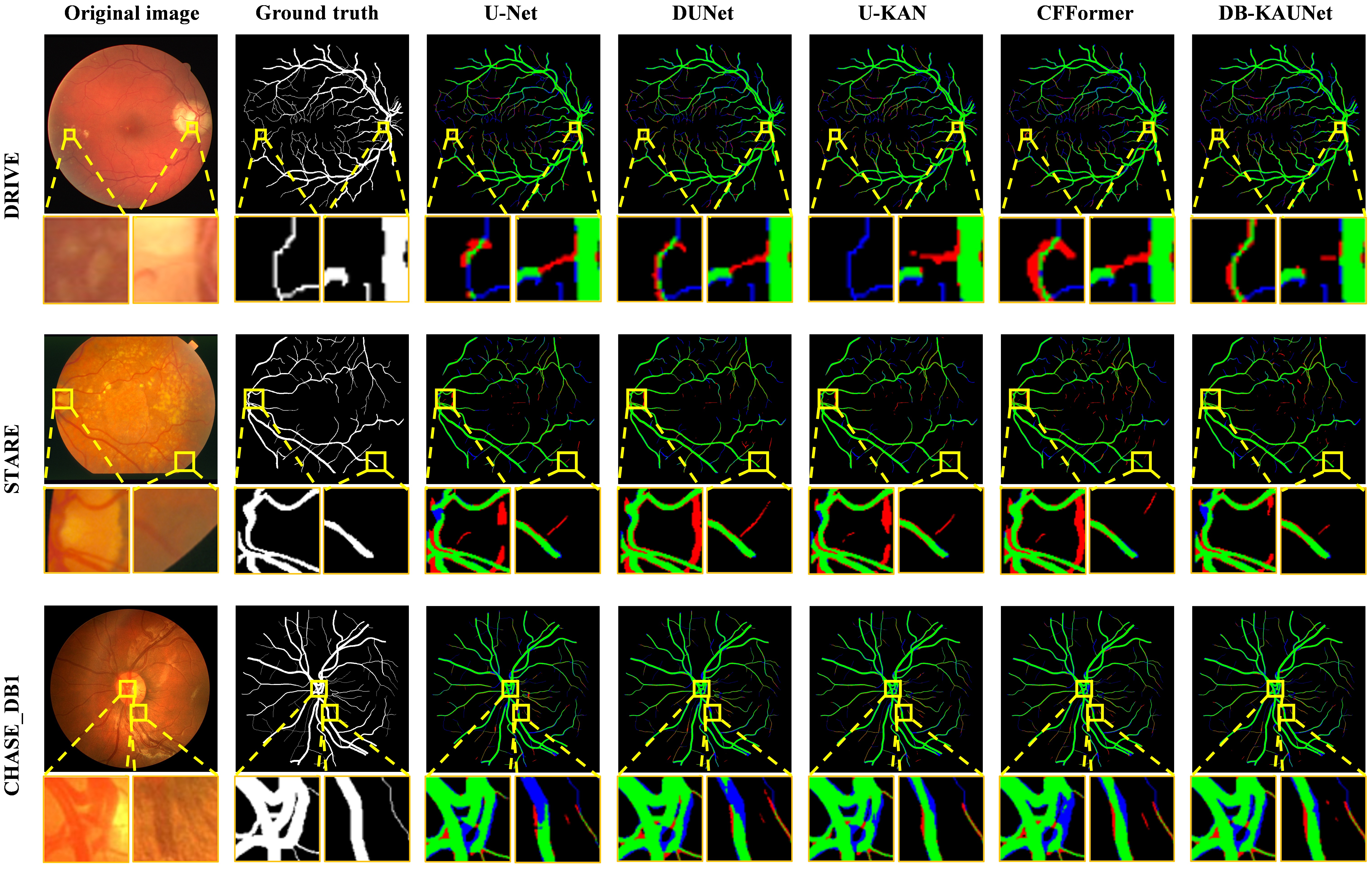} 
    \caption{Qualitative comparison of different segmentation models on the DRIVE, STARE, and CHASE\_DB1 datasets.}
    \label{fig:qualitative_comparison}
\end{figure*}

We compare the DB-KAUNet model with various retinal vessel segmentation methods on the DRIVE, STARE, and CHASE\_DB1 datasets. The experimental results are presented in Figure~\ref{fig:qualitative_comparison} and Tables~\ref{tab:sota_drive}-\ref{tab:sota_chase}. The compared models include U-Net \cite{2015U-Net} and its typical variants, such as Attn U-Net \cite{2018Attention}, R2U-Net \cite{2018R2U-Net}, UNet++ \cite{2018UNet++}, and DUNet \cite{2019DUNet}. We also include recent state-of-the-art methods: Bridge-Net \cite{2022Bridge-Net}, DPF-Net \cite{2023DPF-Net}, RVS-FDSC \cite{2024RVS-FDSC}, U-KAN \cite{2025U-kan}, Mid-Net \cite{2025Mid-Net}, PA-Net \cite{2025PA-Net}, MDF-Net \cite{2026MDF-Net}, SA-UNet \cite{2026SA-UNet},DSFN-Net \cite{2026DSFN-Net}, and CFFormer \cite{2026CFFormer}.

\begin{table}[!htb]
    \centering
    \caption{Quantitative Comparison with State-of-the-Art Methods on the DRIVE Dataset. Best results are in \textcolor{red}{\textbf{red}}, second best in \textcolor{blue}{blue}.}
    \label{tab:sota_drive}
    \resizebox{\columnwidth}{!}{%
    \begin{tabular}{lcccccc}
        \toprule
        \textbf{Method} & \textbf{Year} & \textbf{AUC} & \textbf{F1} & \textbf{SE} & \textbf{SP} & \textbf{ACC} \\
        \midrule
        U-Net \cite{2015U-Net} & 2015 & 0.9766 & 0.8108 & 0.7776 & 0.9867 & 0.9681 \\
        Attn U-Net \cite{2018Attention} & 2018 & 0.9813 & 0.8299 & 0.8187 & 0.9775 & 0.9573 \\
        R2U-Net \cite{2018R2U-Net} & 2018 & 0.9784 & 0.8171 & 0.7792 & 0.9813 & 0.9636 \\
        UNet++ \cite{2018UNet++} & 2018 & 0.9750 & 0.8111 & 0.8031 & 0.9820 & 0.9533 \\
        DUNet \cite{2019DUNet} & 2019 & 0.9856 & 0.8203 & 0.7894 & 0.9870 & 0.9697 \\
        Bridge-Net \cite{2022Bridge-Net} & 2022 & 0.9834 & 0.8203 & 0.7853 & 0.9818 & 0.9565 \\
        DPF-Net \cite{2023DPF-Net} & 2023 & 0.9824 & 0.8303 & 0.8279 & 0.9776 & 0.9570 \\
        RVS-FDSC \cite{2024RVS-FDSC} & 2024 & 0.9856 & 0.8271 & 0.8221 & \secondbest{0.9878} & 0.9692 \\
        U-KAN \cite{2025U-kan} & 2025 & \secondbest{0.9930} & 0.8871 & \secondbest{0.8950} & 0.9824 & 0.9714 \\
        Mid-Net \cite{2025Mid-Net} & 2025 & 0.9864 & 0.8526 & 0.8721 & 0.9759 & 0.9630 \\
        PA-Net \cite{2025PA-Net} &2025 & 0.9833 & 0.8393 & 0.8284 & 0.9807 & 0.9582 \\
        MDF-Net \cite{2026MDF-Net} & 2026 & 0.9853 & 0.8404 & 0.8505 & 0.9787 & 0.9631 \\
        SA-UNet \cite{2026SA-UNet} & 2026 & 0.9881 & 0.8182 & 0.7842 & \best{0.9971} & 0.9701 \\
        DSFN-Net \cite{2026DSFN-Net} & 2026 & 0.9883 & 0.8321 & 0.8221 & 0.9856 & 0.9711 \\
        CFFormer \cite{2026CFFormer} & 2026 & 0.9913 & \secondbest{0.8916} & 0.8894 & 0.9845 & \secondbest{0.9723} \\
        \midrule
        \textbf{DB-KAUNet (Ours)} & - & \best{0.9937} & \best{0.8964} & \best{0.8985} & 0.9848 & \best{0.9739} \\
        \bottomrule
    \end{tabular}%
    }
\end{table}

We first perform a quantitative evaluation of the experimental results on the DRIVE dataset. As shown in Table~\ref{tab:sota_drive}, our proposed DB-KAUNet achieves the best performance on four key metrics. Notably, the model obtains an AUC of $0.9937$, which indicates its strong discriminative ability in distinguishing between vessel and non-vessel pixels. Concurrently, a high F1 of $0.8964$ reflects an excellent balance between precision and recall. Furthermore, a SE of $0.8985$ demonstrates its proficiency in correctly identifying true vessel pixels, while an ACC of $0.9739$ confirms its strong overall segmentation correctness. Although the SP of DB-KAUNet ($0.9848$) is not the highest, it remains highly competitive. In contrast, while SA-UNet possesses the highest SP, its SE is $0.7842$. This notable gap indicates an imbalanced classification performance, which suggests a difficulty in fully addressing the inherent challenge of class imbalance in retinal vessel segmentation. Complementing these quantitative metrics, the qualitative results in Figure~\ref{fig:qualitative_comparison} further showcase our model's advantages. DB-KAUNet can effectively suppress interference from exudates, thus avoiding the misclassification of this noise as vessels. More importantly, our method is also capable of performing more precise segmentation, even in overexposed regions of the image.

\begin{table}[!htb]
    \centering
    \caption{Quantitative Comparison with State-of-the-Art Methods on the STARE Dataset. Best results are in \textcolor{red}{\textbf{red}}, second best in \textcolor{blue}{blue}.}
    \label{tab:sota_stare}
    \resizebox{\columnwidth}{!}{%
    \begin{tabular}{lcccccc}
        \toprule
        \textbf{Method} & \textbf{Year} & \textbf{AUC} & \textbf{F1} & \textbf{SE} & \textbf{SP} & \textbf{ACC} \\
        \midrule
        U-Net \cite{2015U-Net} & 2015 & 0.9899 & 0.8421 & 0.8135 & 0.9883 & 0.9728 \\
        Attn U-Net \cite{2018Attention} & 2018 & 0.9850 & 0.8357 & 0.8044 & 0.9850 & 0.9650 \\
        R2U-Net \cite{2018R2U-Net} & 2018 & 0.9874 & 0.8433 & 0.8245 & 0.9832 & 0.9722 \\
        UNet++ \cite{2018UNet++} & 2018 & 0.9822 & 0.8322 & 0.8223 & 0.9848 & 0.9686 \\
        DUNet \cite{2019DUNet} & 2019 & 0.9868 & 0.8079 & 0.7428 & \best{0.9920} & 0.9729 \\
        Bridge-Net \cite{2022Bridge-Net} & 2022 & 0.9901 & 0.8289 & 0.8002 & 0.9864 & 0.9668 \\
        DPF-Net \cite{2023DPF-Net} & 2023 & 0.9854 & 0.8366 & 0.8287 & 0.9854 & 0.9655 \\
        RVS-FDSC \cite{2024RVS-FDSC} & 2024 & 0.9808 & 0.7875 & 0.8004 & 0.9882 & 0.9718 \\
        U-KAN \cite{2025U-kan} & 2025 & \secondbest{0.9967} & 0.9132 & \best{0.9230} & 0.9884 & 0.9815 \\
        Mid-Net \cite{2025Mid-Net} & 2025 & 0.9923 & 0.8500 & 0.8787 & 0.9842 & 0.9762 \\
        PA-Net \cite{2025PA-Net} &2025 & 0.9908 & 0.8561 & 0.8813 & 0.9805 & 0.9709 \\
        MDF-Net \cite{2026MDF-Net} & 2026 & 0.9936 & 0.8703 & 0.8445 & 0.9888 & 0.9737 \\
        SA-UNet \cite{2026SA-UNet} & 2026 & 0.9671 & 0.7305 & 0.6593 & 0.9844 & 0.9604 \\
        DSFN-Net \cite{2026DSFN-Net} & 2026 & 0.9890 & 0.8478 & 0.8707 & 0.9852 & 0.9766 \\
        CFFormer \cite{2026CFFormer} & 2026 & \best{0.9969} & \secondbest{0.9163} & 0.9159 & \secondbest{0.9902} & \secondbest{0.9824} \\
        \midrule
        \textbf{DB-KAUNet (Ours)} & - & \best{0.9969} & \best{0.9183} & \secondbest{0.9208} & 0.9900 & \best{0.9828} \\
        \bottomrule
    \end{tabular}%
    }
\end{table}

For the quantitative analysis on the STARE dataset, as shown in Table~\ref{tab:sota_stare}, our DB-KAUNet demonstrates its superiority across three key performance metrics. The model achieves an AUC of $0.9969$, highlighting its excellent capability in distinguishing between positive and negative samples. Additionally, a high F1 of $0.9183$ and an ACC of $0.9828$ demonstrate its exceptional balance between precision and recall, as well as its strong overall segmentation accuracy. Notably, the model not only obtains the second-best result for the SE metric, but its SP value is also significantly superior to most comparative methods, which collectively reflects the robustness of its performance. This strong quantitative performance can be attributed to the model's ability to overcome common segmentation challenges, a point that is visually confirmed by the qualitative results in Figure~\ref{fig:qualitative_comparison}. Through visual comparison, it is evident that DB-KAUNet can more effectively mitigate the adverse effects of the optic disc region on the segmentation results compared to several other models. This clearly indicates that our method possesses higher precision in differentiating true vessels from the complex background.

\begin{table}[!htb]
    \centering
    \caption{Quantitative Comparison with State-of-the-Art Methods on the CHASE\_DB1 Dataset. Best results are in \textcolor{red}{\textbf{red}}, second best in \textcolor{blue}{blue}.}
    \label{tab:sota_chase}
    \resizebox{\columnwidth}{!}{%
    \begin{tabular}{lcccccc}
        \toprule
        \textbf{Method} & \textbf{Year} & \textbf{AUC} & \textbf{F1} & \textbf{SE} & \textbf{SP} & \textbf{ACC} \\
        \midrule
        U-Net \cite{2015U-Net} & 2015 & 0.9881 & 0.8026 & 0.8146 & 0.9855 & 0.9747 \\
        Attn U-Net \cite{2018Attention} & 2018 & 0.9842 & 0.8200 & 0.8130 & 0.9818 & 0.9655 \\
        R2U-Net \cite{2018R2U-Net} & 2018 & 0.9815 & 0.7928 & 0.7756 & 0.9820 & 0.9635 \\
        UNet++ \cite{2018UNet++} & 2018 & 0.9781 & 0.8203 & 0.8133 & 0.9809 & 0.9610 \\
        DUNet \cite{2019DUNet} & 2019 & 0.9863 & 0.7853 & 0.8229 & 0.9821 & 0.9724 \\
        Bridge-Net \cite{2022Bridge-Net} & 2022 & 0.9893 & 0.8293 & 0.8132 & 0.9840 & 0.9667 \\
        DPF-Net \cite{2023DPF-Net} & 2023 & 0.9868 & 0.8302 & 0.8303 & 0.9841 & 0.9676 \\
        RVS-FDSC \cite{2024RVS-FDSC} & 2024 & 0.9867 & 0.8050 & 0.8356 & 0.9856 & 0.9743 \\
        U-KAN \cite{2025U-kan} & 2025 & \secondbest{0.9916} & \secondbest{0.8656} & 0.8723 & 0.9847 & 0.9738 \\
        Mid-Net \cite{2025Mid-Net} & 2025 & \best{0.9926} & 0.8550 & 0.8673 & 0.9866 & \best{0.9774} \\
        PA-Net \cite{2025PA-Net} &2025 & 0.9875 & 0.8308 & 0.8570 & 0.9779 & 0.9677 \\
        MDF-Net \cite{2026MDF-Net} & 2026 & 0.9903 & 0.8420 & 0.7843 & \secondbest{0.9903} & 0.9676 \\
        SA-UNet \cite{2026SA-UNet} & 2026 & 0.9864 & 0.8218 & 0.8102 & \best{0.9936} & 0.9696 \\
        DSFN-Net \cite{2026DSFN-Net} & 2026 & 0.9913 & 0.8206 & 0.8580 & 0.9845 & \secondbest{0.9764} \\
        CFFormer \cite{2026CFFormer} & 2026 & 0.9914 & 0.8652 & \secondbest{0.8726} & 0.9846 & 0.9738 \\
        \midrule
        \textbf{DB-KAUNet (Ours)} & - & \secondbest{0.9916} & \best{0.8678} & \best{0.8750} & 0.9849 & 0.9743 \\
        \bottomrule
    \end{tabular}%
    }
\end{table}

The quantitative analysis of the CHASE\_DB1 dataset further confirms the competitiveness of our method. As presented in Table~\ref{tab:sota_chase}, DB-KAUNet achieves top performance on two key metrics, with an F1 of $0.8678$ and a SE of $0.8750$. At the same time, its AUC value of $0.9916$ ranks as the second-highest, which collectively validates the model's superior performance. Although some comparative methods attain higher values for SP and ACC, DB-KAUNet demonstrates a more balanced and comprehensive capability across all evaluation dimensions. This is a crucial attribute for processing complex medical images. This performance advantage is substantiated visually by the qualitative results in Figure~\ref{fig:qualitative_comparison}. In low-contrast regions caused by high exposure, many other models exhibit significant vessel omissions. In sharp contrast, our method can meticulously capture and completely delineate these difficult-to-discern target vessels, powerfully demonstrating its robustness and precision under adverse imaging conditions.

\subsection{Complexity Analysis}
\label{subsec:Complexity analysis}

\begin{table}[!htb]
    \centering
    \caption{Complexity Analysis of Different Models. Best results are in \textcolor{red}{\textbf{red}}, second best in \textcolor{blue}{blue}.}
    \label{tab:complexity}
    \resizebox{\columnwidth}{!}{
    \begin{tabular}{lcccc}
        \toprule
        \textbf{Method} & \textbf{Parameters (M) $\downarrow$} & \textbf{FLOPs (G) $\downarrow$} & \textbf{F1(\%) $\uparrow$} \\
        \midrule
        U-Net \cite{2015U-Net} & 34.53 & 4.09 & 81.08 \\
        Attn U-Net \cite{2018Attention} & 34.88 & 50.96 & 82.99 \\
        U-KAN \cite{2025U-kan} & \secondbest{25.36} & \best{0.43} & 88.71 \\
        DSFN-Net \cite{2026DSFN-Net} & \best{23.40} & 94.78 & 83.21 \\
        CFFormer \cite{2026CFFormer} & 100.54 & 3.62 & \secondbest{89.16} \\
        DB-KAUNet (Ours) & 96.31 & \secondbest{1.72} & \best{89.64} \\
        \bottomrule
    \end{tabular}}
\end{table}

To evaluate the overall effectiveness of our model, we compare the computational complexity and segmentation performance of DB-KAUNet against five other architectures. The analysis aims to illustrate the trade-off between performance and efficiency. As detailed in Table \ref{tab:complexity}, we examine the parameters, Floating Point Operations (FLOPs), and F1 for each model.

The analysis reveals complex trade-offs among the different models. U-Net \cite{2015U-Net} serves as the primary baseline, offering modest accuracy (81.08\% F1) with a relatively high computational load (4.09 G FLOPs). Other models present varied strategies: Attn U-Net \cite{2018Attention} slightly improves accuracy but at the cost of an exceptionally high 50.96 G FLOPs. DSFN-Net \cite{2026DSFN-Net} achieves the lowest parameter count (23.40 M) but incurs the highest computational load (94.78 G). In sharp contrast, U-KAN \cite{2025U-kan} showcases the theoretical efficiency of KANs, delivering robust segmentation (88.71\% F1) with a remarkably low computational cost of only 0.43 G FLOPs. At the other extreme, CFFormer \cite{2026CFFormer} represents a high-cost, high-performance model, using the most parameters (100.54 M) to achieve the second-best F1 (89.16\%).

Against this backdrop, our proposed DB-KAUNet presents a compelling trade-off between performance and efficiency. It achieves the highest F1 of 89.64\%, establishing its superior segmentation accuracy among the compared models. Although DB-KAUNet has a large parameter count (96.31 M), its computational core is extremely efficient. The model requires only 1.72 G FLOPs, the second-best figure in the comparison. Achieving such a low computational load despite a large parameter size highlights the operational efficiency of our proposed KAN-based components.

In summary, the advantage of DB-KAUNet is its ability to deliver state-of-the-art performance while maintaining highly competitive computational efficiency in terms of FLOPs. This makes it an advanced solution that achieves an ideal equilibrium between the pursuit of high accuracy and the constraints of computational resources.

\section{Conclusion}
\label{sec:Conclusion}
In this paper, we propose a novel network named DB-KAUNet to achieve high precision in retinal vessel segmentation. The core of this network is the HDBE, which is responsible for comprehensively capturing local details, long-range dependencies, and complex nonlinear associations. The CCI module ensures efficient information fusion between the dual encoder channels. Furthermore, the innovative SFE and SFE-GAF modules further optimize spatial feature fusion and sampling efficiency. Comprehensive evaluations on the DRIVE, STARE, and CHASE\_DB1 datasets validate the effectiveness of the proposed method. The experimental results show that DB-KAUNet not only surpasses existing methods across various metrics but also demonstrates excellent segmentation performance and robustness.

\section*{Funding sources}
Funding: This work was supported by the National Natural Science Foundation of China [grant numbers 62502161, 51365017]; and the Natural Science Foundation of Jiangxi Province of China [grant number 20192BAB205084].

\section*{CRediT authorship contribution statement}
\textbf{Hongyu Xu:} Writing – original draft, Visualization, Software, Methodology, Investigation, Conceptualization. \textbf{Panpan Meng:} Writing – review \& editing, Formal analysis, Data curation. \textbf{Meng Wang:} Writing – review \& editing, Resources, Conceptualization. \textbf{Dayu Hu:} Validation, Supervision. \textbf{Liming Liang:} Supervision, Conceptualization. \textbf{Xiaoqi Sheng:} Writing – review \& editing, Supervision, Methodology, Investigation, Funding acquisition.

\section*{Declaration of competing interest}
The authors declare that they have no known competing financial interests or personal relationships that could have appeared to influence the work reported in this paper.

\normalsize
\bibliography{references}

@article{2017Retinal,
  title={{Retinal vasculature in glaucoma: A review}},
  author={Chan, Karen K W and Tang, Fangyao and Tham, Clement C Y and Young, Alvin L and Cheung, Carol Y},
  journal={BMJ Open Ophthalmol.},
  volume={1},
  number={1},
  pages={e000032},
  doi = {10.1136/bmjophth-2016-000032},
  year={2017},
}

@article{Carol2011Retinal,
title = {{Retinal Vascular Tortuosity, Blood Pressure, and Cardiovascular Risk Factors}},
journal = {Ophthalmol.},
volume = {118},
number = {5},
pages = {812-818},
year = {2011},
issn = {0161-6420},
doi = {10.1016/j.ophtha.2010.08.045},
author = {Carol Yim-lui Cheung and Yingfeng Zheng and Wynne Hsu and Mong Li Lee and Qiangfeng Peter Lau and Paul Mitchell and Jie Jin Wang and Ronald Klein and Tien Yin Wong}
}

@INPROCEEDINGS{2014Classification,
  author={Irshad, Samra and Akram, M. Usman},
  booktitle={2014 Cairo International Biomedical Engineering Conference (CIBEC)}, 
  title={{Classification of retinal vessels into arteries and veins for detection of hypertensive retinopathy}}, 
  year={2014},
  volume={},
  number={},
  pages={133-136},
  doi={10.1109/CIBEC.2014.7020937}
}

@misc{2018Attention,
title={{Attention U-Net: Learning Where to Look for the Pancreas}}, 
author={Ozan Oktay and Jo Schlemper and Loic Le Folgoc and Matthew Lee and Mattias Heinrich and Kazunari Misawa and Kensaku Mori and Steven McDonagh and Nils Y Hammerla and Bernhard Kainz and Ben Glocker and Daniel Rueckert},
year={2018},
eprint={1804.03999},
archivePrefix={arXiv},
primaryClass={cs.CV}
}

@InProceedings{2018UNet++,
author={Zhou, Zongwei
and Rahman Siddiquee, Md Mahfuzur
and Tajbakhsh, Nima
and Liang, Jianming},
title={{UNet++: A Nested U-Net Architecture for Medical Image Segmentation}},
booktitle={Deep Learning in Medical Image Analysis and Multimodal Learning for Clinical Decision Support},
year={2018},
publisher={Springer International Publishing},
address={Cham},
pages={3--11},
doi={10.1007/978-3-030-00889-5_1}
}

@InProceedings{2015U-Net,
author={Ronneberger, Olaf
and Fischer, Philipp
and Brox, Thomas},
title={{U-Net: Convolutional Networks for Biomedical Image Segmentation}},
booktitle={Medical Image Computing and Computer-Assisted Intervention -- MICCAI 2015},
year={2015},
publisher={Springer International Publishing},
address={Cham},
pages={234--241},
doi={10.1007/978-3-319-24574-4_28}
}

@INPROCEEDINGS{2018R2U-Net,
  author={Alom, Md Zahangir and Yakopcic, Chris and Taha, Tarek M. and Asari, Vijayan K.},
  booktitle={NAECON 2018 - IEEE National Aerospace and Electronics Conference}, 
  title={{Nuclei Segmentation with Recurrent Residual Convolutional Neural Networks based U-Net (R2U-Net)}}, 
  year={2018},
  volume={},
  number={},
  pages={228-233},
  doi={10.1109/NAECON.2018.8556686}}

@article{2019DUNet,
title = {{DUNet: A deformable network for retinal vessel segmentation}},
journal = {Knowl.-Based Syst.},
volume = {178},
pages = {149-162},
year = {2019},
issn = {0950-7051},
doi = {10.1016/j.knosys.2019.04.025},
author = {Qiangguo Jin and Zhaopeng Meng and Tuan D. Pham and Qi Chen and Leyi Wei and Ran Su}
}

@INPROCEEDINGS{2017DCN1,
  author={Dai, Jifeng and Qi, Haozhi and Xiong, Yuwen and Li, Yi and Zhang, Guodong and Hu, Han and Wei, Yichen},
  booktitle={2017 IEEE International Conference on Computer Vision (ICCV)}, 
  title={{Deformable Convolutional Networks}}, 
  year={2017},
  volume={},
  number={},
  pages={764-773},
  doi={10.1109/ICCV.2017.89}}

@inproceedings{2020DCN2,
  author={Zhu, Xizhou and Hu, Han and Lin, Stephen and Dai, Jifeng},
  booktitle={2019 IEEE/CVF Conference on Computer Vision and Pattern Recognition (CVPR)}, 
  title={{Deformable ConvNets V2: More Deformable, Better Results}}, 
  year={2019},
  volume={},
  number={},
  pages={9300-9308},
  doi={10.1109/CVPR.2019.00953}
}

@inproceedings{wang2023DCN3,
  author={Wang, Wenhai and Dai, Jifeng and Chen, Zhe and Huang, Zhenhang and Li, Zhiqi and Zhu, Xizhou and Hu, Xiaowei and Lu, Tong and Lu, Lewei and Li, Hongsheng and Wang, Xiaogang and Qiao, Yu},
  booktitle={2023 IEEE/CVF Conference on Computer Vision and Pattern Recognition (CVPR)}, 
  title={{InternImage: Exploring Large-Scale Vision Foundation Models with Deformable Convolutions}}, 
  year={2023},
  volume={},
  number={},
  pages={14408-14419},
  doi={10.1109/CVPR52729.2023.01385}
}

@inproceedings{2024DCN4,
  author={Xiong, Yuwen and Li, Zhiqi and Chen, Yuntao and Wang, Feng and Zhu, Xizhou and Luo, Jiapeng and Wang, Wenhai and Lu, Tong and Li, Hongsheng and Qiao, Yu and Lu, Lewei and Zhou, Jie and Dai, Jifeng},
  booktitle={2024 IEEE/CVF Conference on Computer Vision and Pattern Recognition (CVPR)}, 
  title={{Efficient Deformable ConvNets: Rethinking Dynamic and Sparse Operator for Vision Applications}}, 
  year={2024},
  volume={},
  number={},
  pages={5652-5661},
  doi={10.1109/CVPR52733.2024.00540}
}

@article{2023LDConv,
title = {{LDConv: Linear deformable convolution for improving convolutional neural networks}},
journal = {Image Vis. Comput.},
volume = {149},
pages = {105190},
year = {2024},
issn = {0262-8856},
doi = {10.1016/j.imavis.2024.105190},
author = {Xin Zhang and Yingze Song and Tingting Song and Degang Yang and Yichen Ye and Jie Zhou and Liming Zhang}
}

@article{2004DRIVE,
  author={Staal, J. and Abramoff, M.D. and Niemeijer, M. and Viergever, M.A. and van Ginneken, B.},
  journal={IEEE Trans. Med. Imaging}, 
  title={{Ridge-based vessel segmentation in color images of the retina}}, 
  year={2004},
  volume={23},
  number={4},
  pages={501-509},
  doi={10.1109/TMI.2004.825627}
}

@article{Hoover2000STARE,
  author={Hoover, A.D. and Kouznetsova, V. and Goldbaum, M.},
  journal={IEEE Trans. Med. Imaging}, 
  title={{Locating blood vessels in retinal images by piecewise threshold probing of a matched filter response}}, 
  year={2000},
  volume={19},
  number={3},
  pages={203-210},
  doi={10.1109/42.845178}
}

@article{2009CHASE,
author = {Owen, Christopher and Rudnicka, Alicja and Mullen, Robert and Barman, Sarah and Monekosso, Dorothy and Whincup, Peter and Ng, Jeffrey and Paterson, Carl},
year = {2009},
month = {04},
pages = {2004-10},
title = {{Measuring Retinal Vessel Tortuosity in 10-Year-Old Children: Validation of the Computer-Assisted Image Analysis of the Retina (CAIAR) Program}},
volume = {50},
journal = {Invest. Ophthalmol. Vis. Sci.},
doi = {10.1167/iovs.08-3018}
}

@inproceedings{2020Squeeze-and-attention,
  author={Zhong, Zilong and Lin, Zhong Qiu and Bidart, Rene and Hu, Xiaodan and Daya, Ibrahim Ben and Li, Zhifeng and Zheng, Wei-Shi and Li, Jonathan and Wong, Alexander},
  booktitle={2020 IEEE/CVF Conference on Computer Vision and Pattern Recognition (CVPR)}, 
  title={{Squeeze-and-Attention Networks for Semantic Segmentation}}, 
  year={2020},
  volume={},
  number={},
  pages={13062-13071},
  doi={10.1109/CVPR42600.2020.01308}
}

@misc{liu2024kAN,
      title={{KAN: Kolmogorov-Arnold Networks}}, 
      author={Ziming Liu and Yixuan Wang and Sachin Vaidya and Fabian Ruehle and James Halverson and Marin Soljačić and Thomas Y. Hou and Max Tegmark},
      year={2025},
      eprint={2404.19756},
      archivePrefix={arXiv},
      primaryClass={cs.LG}
}

@INPROCEEDINGS{2016ResNet,
  author={He, Kaiming and Zhang, Xiangyu and Ren, Shaoqing and Sun, Jian},
  booktitle={2016 IEEE Conference on Computer Vision and Pattern Recognition (CVPR)}, 
  title={{Deep Residual Learning for Image Recognition}}, 
  year={2016},
  volume={},
  number={},
  pages={770-778},
  doi={10.1109/CVPR.2016.90}
}

@ARTICLE{2022DWConv,
  author={Cao, Jinming and Li, Yangyan and Sun, Mingchao and Chen, Ying and Lischinski, Dani and Cohen-Or, Daniel and Chen, Baoquan and Tu, Changhe},
  journal={IEEE Trans. Image Process.}, 
  title={{DO-Conv: Depthwise Over-Parameterized Convolutional Layer}}, 
  year={2022},
  volume={31},
  number={},
  pages={3726-3736},
  doi={10.1109/TIP.2022.3175432}
}

@article{2018SiLU,
title = {{Sigmoid-weighted linear units for neural network function approximation in reinforcement learning}},
journal = {Neural Netw.},
volume = {107},
pages = {3-11},
year = {2018},
issn = {0893-6080},
doi = {10.1016/j.neunet.2017.12.012},
author = {Stefan Elfwing and Eiji Uchibe and Kenji Doya},
}

@article{2025TAOD-CFNet,
title = {{A precise image-based retinal blood vessel segmentation method using TAOD-CFNet}},
journal = {Biomed. Signal Process. Control},
volume = {107},
pages = {107815},
year = {2025},
issn = {1746-8094},
doi = {10.1016/j.bspc.2025.107815},
author = {Yixin Yang and Lixiang Sun and Zhiwen Tang and Genhua Liu and Guoxiong Zhou and Lin Li and Weiwei Cai and Liujun Li and Lin Chen and Linan Hu},
}

@article{2023HCTNet,
title = {{HCTNet: A hybrid CNN-transformer network for breast ultrasound image segmentation}},
journal = {Comput. Biol. Med.},
volume = {155},
pages = {106629},
year = {2023},
issn = {0010-4825},
doi = {10.1016/j.compbiomed.2023.106629},
author = {Qiqi He and Qiuju Yang and Minghao Xie}
}

@article{2023SGAT-Net,
title = {{Stimulus-guided adaptive transformer network for retinal blood vessel segmentation in fundus images}},
journal = {Med. Image Anal.},
volume = {89},
pages = {102929},
year = {2023},
issn = {1361-8415},
doi = {10.1016/j.media.2023.102929},
author = {Ji Lin and Xingru Huang and Huiyu Zhou and Yaqi Wang and Qianni Zhang}
}

@article{2023ARP-Net,
title = {{Transformer and convolutional based dual branch network for retinal vessel segmentation in OCTA images}},
journal = {Biomed. Signal Process. Control},
volume = {83},
pages = {104604},
year = {2023},
issn = {1746-8094},
doi = {10.1016/j.bspc.2023.104604},
author = {Xiaoming Liu and Di Zhang and Junping Yao and Jinshan Tang}
}

@article{2024TransUNet,
title = {{TransUNet: Rethinking the U-Net architecture design for medical image segmentation through the lens of transformers}},
journal = {Med. Image Anal.},
volume = {97},
pages = {103280},
year = {2024},
issn = {1361-8415},
doi = {10.1016/j.media.2024.103280},
author = {Jieneng Chen and Jieru Mei and Xianhang Li and Yongyi Lu and Qihang Yu and Qingyue Wei and Xiangde Luo and Yutong Xie and Ehsan Adeli and Yan Wang and Matthew P. Lungren and Shaoting Zhang and Lei Xing and Le Lu and Alan Yuille and Yuyin Zhou}
}

@article{2026CFFormer,
title = {{CFFormer: Cross CNN-Transformer channel attention and spatial feature fusion for improved segmentation of heterogeneous medical images}},
journal = {Expert Syst. Appl.},
volume = {295},
pages = {128835},
year = {2026},
issn = {0957-4174},
doi = {10.1016/j.eswa.2025.128835},
author = {Jiaxuan Li and Qing Xu and Xiangjian He and Ziyu Liu and Daokun Zhang and Ruili Wang and Rong Qu and Guoping Qiu}
}

@inproceedings{2025U-kan,
author = {Li, Chenxin and Liu, Xinyu and Li, Wuyang and Wang, Cheng and Liu, Hengyu and Liu, Yifan and Chen, Zhen and Yuan, Yixuan},
title = {{U-KAN makes strong backbone for medical image segmentation and generation}},
year = {2025},
publisher = {AAAI Press},
doi = {10.1609/aaai.v39i5.32491},
booktitle = {Proceedings of the AAAI Conference on Artificial Intelligence},
pages={4652--4660}
}

@article{2026MDF-Net,
title = {{MDF-Net: An attention-guided multi-scale dual-fusion network for retinal vessel segmentation}},
journal = {Meas.},
volume = {257},
pages = {118695},
year = {2026},
issn = {0263-2241},
doi = {10.1016/j.measurement.2025.118695},
author = {Mingjun Ma and Liming Liang and Xiaoqi Sheng}
}

@article{2026DSFN-Net,
title = {{A Dual-Layer Semantic Fusion Network for Retinal Vessel Segmentation}},
journal = {Biomed. Signal Process. Control},
volume = {112},
pages = {108463},
year = {2026},
issn = {1746-8094},
doi = {10.1016/j.bspc.2025.108463},
author = {Zizheng Li and Huadeng Wang and Ningning Tang and Rushi Lan and Bo Li and Xiaonan Luo}
}

@article{2024RVS-FDSC,
title = {{RVS-FDSC: A retinal vessel segmentation method with four-directional strip convolution to enhance feature extraction}},
journal = {Biomed. Signal Process. Control},
volume = {95},
pages = {106296},
year = {2024},
issn = {1746-8094},
doi = {10.1016/j.bspc.2024.106296},
author = {Linfeng Kong and Yun Wu}
}

@article{2022Bridge-Net,
title = {{Bridge-Net: Context-involved U-net with patch-based loss weight mapping for retinal blood vessel segmentation}},
journal = {Expert Syst. Appl.},
volume = {195},
pages = {116526},
year = {2022},
issn = {0957-4174},
doi = {10.1016/j.eswa.2022.116526},
author = {Yuan Zhang and Miao He and Zhineng Chen and Kai Hu and Xuanya Li and Xieping Gao}
}

@article{2025Mid-Net,
title = {{Mid-Net: Rethinking efficient network architectures for small-sample vascular segmentation}},
journal = {Inf. Fusion},
volume = {115},
pages = {102777},
year = {2025},
issn = {1566-2535},
doi = {10.1016/j.inffus.2024.102777},
author = {Dongxin Zhao and Jianhua Liu and Peng Geng and Jiaxin Yang and Ziqian Zhang and Yin Zhang},
}

@ARTICLE{2023DPF-Net,
  author={Li, Jianyong and Gao, Ge and Yang, Lei and Bian, Guibin and Liu, Yanhong},
  journal={IEEE Trans. Instrum. Meas.}, 
  title={{DPF-Net: A Dual-Path Progressive Fusion Network for Retinal Vessel Segmentation}}, 
  year={2023},
  volume={72},
  number={},
  pages={1-17},
  doi={10.1109/TIM.2023.3277946}
}

@article{2026SA-UNet,
title = {{Multimodal self-supervised retinal vessel segmentation}},
journal = {Neural Netw.},
volume = {193},
pages = {108011},
year = {2026},
issn = {0893-6080},
doi = {10.1016/j.neunet.2025.108011},
author = {Pengshuai Yin and Jingqi Zhang and Huichou Huang and Ruirui Liu and Yanxia Liu and Qingyao Wu and F. Richard Yu}
}

@INPROCEEDINGS{2019PAM,
  author={Fu, Jun and Liu, Jing and Tian, Haijie and Li, Yong and Bao, Yongjun and Fang, Zhiwei and Lu, Hanqing},
  booktitle={2019 IEEE/CVF Conference on Computer Vision and Pattern Recognition (CVPR)}, 
  title={{Dual Attention Network for Scene Segmentation}}, 
  year={2019},
  volume={},
  number={},
  pages={3141-3149},
  doi={10.1109/CVPR.2019.00326}}

@INPROCEEDINGS{2016Dice,
  author={Milletari, Fausto and Navab, Nassir and Ahmadi, Seyed-Ahmad},
  booktitle={2016 Fourth International Conference on 3D Vision (3DV)}, 
  title={{V-Net: Fully Convolutional Neural Networks for Volumetric Medical Image Segmentation}}, 
  year={2016},
  volume={},
  number={},
  pages={565-571},
  doi={10.1109/3DV.2016.79}
}

@misc{2024kat,
      title={{Kolmogorov-Arnold Transformer}}, 
      author={Xingyi Yang and Xinchao Wang},
      year={2024},
      eprint={2409.10594},
      archivePrefix={arXiv},
      primaryClass={cs.LG},
}

@misc{2021Vit,
      title={{An Image is Worth 16x16 Words: Transformers for Image Recognition at Scale}}, 
      author={Alexey Dosovitskiy and Lucas Beyer and Alexander Kolesnikov and Dirk Weissenborn and Xiaohua Zhai and Thomas Unterthiner and Mostafa Dehghani and Matthias Minderer and Georg Heigold and Sylvain Gelly and Jakob Uszkoreit and Neil Houlsby},
      year={2021},
      eprint={2010.11929},
      archivePrefix={arXiv},
      primaryClass={cs.CV},
}

@article{2009mode,
author = {Kolda, Tamara and Bader, Brett},
year = {2009},
month = {08},
pages = {455-500},
title = {{Tensor Decompositions and Applications}},
volume = {51},
journal = {SIAM Rev.},
doi = {10.1137/07070111X}
}

@misc{2020Pad,
      title={{Pad\'e Activation Units: End-to-end Learning of Flexible Activation Functions in Deep Networks}}, 
      author={Alejandro Molina and Patrick Schramowski and Kristian Kersting},
      year={2020},
      eprint={1907.06732},
      archivePrefix={arXiv},
      primaryClass={cs.LG},
}

@article{2025PA-Net,
title = {{PA-Net: A hybrid architecture for retinal vessel segmentation}},
journal = {Pattern Recognit.},
volume = {161},
pages = {111254},
year = {2025},
issn = {0031-3203},
doi = {10.1016/j.patcog.2024.111254},
author = {Xuebing Luo and Lingxi Peng and Ziyan Ke and Jinhui Lin and Zhiwen Yu},
}

\end{document}